\documentclass[10pt,twocolumn,letterpaper]{article}

\usepackage[pagenumbers]{cvpr} 

\usepackage[most]{tcolorbox}

\newtcolorbox{disclaimerbox}{
    colback=gray!10,     
    colframe=gray!40,    
    boxrule=0.5pt,       
    arc=4pt,             
    auto outer arc,
    boxsep=5pt,
    left=6pt,
    right=6pt,
    top=4pt,
    bottom=4pt,
    enhanced jigsaw
}

\usepackage{graphicx}
\usepackage{amsmath}
\usepackage{amssymb}
\usepackage{booktabs}
\usepackage{xcolor}
\usepackage{subcaption}
\usepackage{url}

\definecolor{iccvblue}{rgb}{0.21,0.49,0.74}
\usepackage[pagebackref,breaklinks,colorlinks,allcolors=iccvblue]{hyperref}

\def\paperID{*****} 
\def\confName{CVPR}
\def\confYear{2026}

\title{GenSmoke-GS: A Multi-Stage Method for Novel View Synthesis from Smoke-Degraded Images Using a Generative Model}

\author{
Qida Cao$^{1}$ \quad Xinyuan Hu$^{1}$ \quad Changyue Shi$^{1}$ \\
Jiajun Ding$^{1}$ \quad Zhou Yu$^{1}$ \quad Jun Yu$^{2}$ \\
$^{1}$School of Computer Science and Technology, Hangzhou Dianzi University \\
$^{2}$School of Computer Science and Technology, Harbin Institute of Technology, Shenzhen
}

\begin{document}
\maketitle

\begin{abstract}
This paper describes our method for Track 2 of the NTIRE 2026 3D Restoration and Reconstruction (3DRR) Challenge on smoke-degraded images. In this task, smoke reduces image visibility and weakens the cross-view consistency required by scene optimization and rendering. We address this problem with a multi-stage pipeline consisting of image restoration, dehazing, MLLM-based enhancement, 3DGS-MCMC optimization, and averaging over repeated runs. The main purpose of the pipeline is to improve visibility before rendering while limiting scene-content changes across input views. Experimental results on the challenge benchmark show improved quantitative performance and better visual quality than the provided baselines. The code is available at \url{https://github.com/plbbl/GenSmoke-GS}.
\end{abstract}

\begin{disclaimerbox}
Our method achieved a ranking of 1 out of 14 participants in Track 2 of the NTIRE 3DRR Challenge, as reported on the official competition website: \url{https://www.codabench.org/competitions/13993/#/results-tab}.
\end{disclaimerbox}

\section{Introduction}
\label{sec:intro}

Generating clean novel views from smoke-degraded images is difficult because smoke affects image visibility and also affects the consistency required by the optimization and rendering process. In the NTIRE 2026 3D Restoration and Reconstruction (3DRR) Challenge, the goal is to generate clean views from degraded inputs. When the degradation is severe, directly applying an optimization method to the input images often leads to unstable geometry and low-quality rendered results.

A common way to address this problem is to enhance the input images before optimization. However, in this task, better visual quality in individual images does not always lead to better novel view synthesis results. This is especially true for generative enhancement models. Although such models may improve visibility and recover details, they may also introduce structural differences across images. These differences can reduce the stability of the optimization stage and affect the final rendered views.

For this reason, we use a multi-stage method for novel view synthesis from smoke-degraded images. The method includes image restoration, dehazing, MLLM-based enhancement, 3DGS-MCMC optimization, and averaging over repeated runs. The main idea is to improve visibility while keeping the enhanced inputs as consistent as possible across views.

The final system contains five stages: preliminary restoration with ConvIR-UDPNet, DCP-based dehazing, MLLM-based enhancement with structural constraints, 3DGS-MCMC optimization with FasterGS, and averaging over multiple independent runs. Experimental results show clear improvements over the compared baselines in both quantitative metrics and visual comparisons.

\section{Related Work}
\label{sec:related_work}

\textbf{Novel view synthesis in degraded scenes.} Recent work extends radiance-field and Gaussian-splatting methods to challenging imaging conditions such as underwater scenes, adverse lighting, and participating media~\cite{levy2023seathru,liu2025i2nerf,cui2025luminance,zhou2025lita,yang2025seasplat,cao2025gc,hu2026srsplat}. These studies show that rendering quality can be improved by introducing degradation-related modeling or suitable priors. Most of them focus on explicit scene modeling, while our setting also involves image enhancement before optimization.

\textbf{3D Gaussian Splatting.} 3DGS is a common method for efficient novel view synthesis~\cite{kerbl20233d}. Later methods further improve optimization stability and efficiency, including 3DGS-MCMC~\cite{kheradmand20243d} and FasterGS~\cite{hahlbohm2026faster}. In our method, this line of work is used in the optimization stage.

\textbf{Relation to our method.} In this task, the main issue is not only image degradation itself, but also whether the enhanced images remain sufficiently consistent for optimization and rendering. For this reason, we use a simple preprocessing and enhancement stage before the optimization step.

\section{Methods}
\label{subsec:methods}

\subsection{Overview}

\begin{figure}[t]
    \centering
    \includegraphics[width=\linewidth]{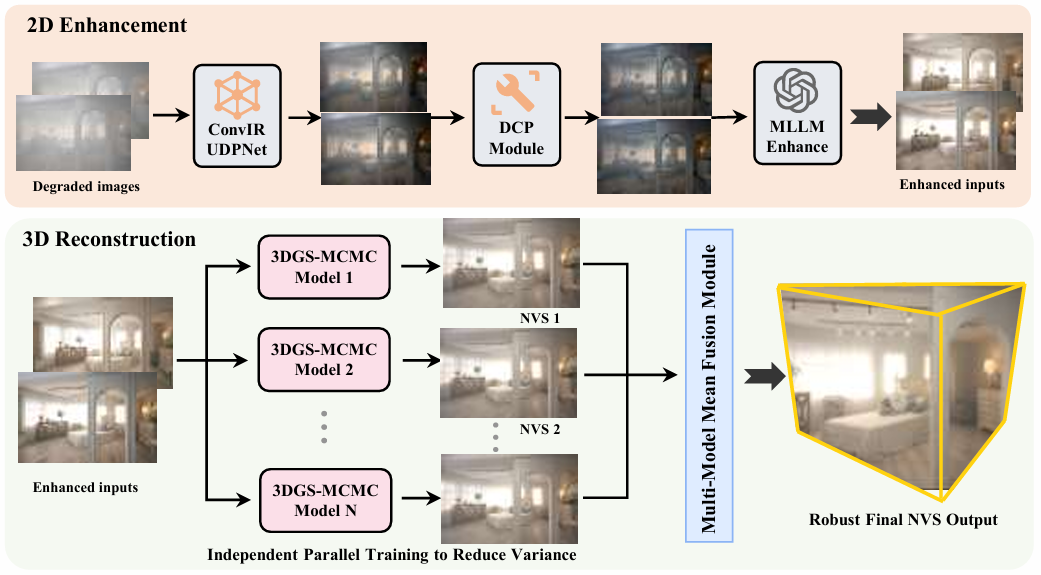}
    \caption{Overview of the method. Smoke-degraded images are first restored and dehazed, then enhanced by an MLLM with structural constraints, and finally processed by 3DGS-MCMC. The final output is obtained by averaging the rendered results of repeated runs.}
    \label{fig:pipeline}
\end{figure}

We use a multi-stage method for novel view synthesis from smoke-degraded images. The goal is to obtain cleaner input images for the optimization stage. As shown in Figure~\ref{fig:pipeline}, the method contains five stages: preliminary restoration, DCP-based dehazing, MLLM-based enhancement, 3DGS-MCMC optimization, and averaging over repeated runs.

\subsection{Preliminary restoration and dehazing}

Given smoke-degraded inputs $\{I_i^{(0)}\}_{i=1}^M$, we first apply ConvIR-UDPNet~\cite{cui2024revitalizing,zuo2026udpnet} to obtain preliminary restored images,
\begin{equation}
    I_i^{(1)} = F_{\text{UDP}}(I_i^{(0)}), \quad i = 1, \ldots, M.
\end{equation}
This step is used to recover coarse structures and basic color information.

We then apply Dark Channel Prior (DCP)~\cite{he2010single} for further dehazing,
\begin{equation}
    I_i^{(2)} = F_{\text{DCP}}(I_i^{(1)}), \quad i = 1, \ldots, M.
\end{equation}
These two steps are used to reduce the influence of heavy degradation before the enhancement stage.

\subsection{MLLM-based enhancement}

After preprocessing, each image is enhanced by an MLLM,
\begin{equation}
    I_i^{(3)} = F_{\text{MLLM}}(I_i^{(2)}, p), \quad i = 1, \ldots, M,
\end{equation}
where $p$ is a prompt used to limit undesired structural changes. We use GPT-Image-1.5~\cite{openai2025gpt} in this stage.

The main issue of this stage is that a generative model may improve visibility while also changing scene content across images. Such changes can affect the following optimization stage. To reduce this problem, we enhance each image independently and use a prompt that asks the model to preserve geometry, layout, object boundaries, and local structures as much as possible, while allowing visibility improvement, denoising, and moderate detail recovery.

\subsection{Optimization and averaging}

The enhanced images are then used for scene optimization with 3DGS-MCMC~\cite{kheradmand20243d}, accelerated by FasterGS~\cite{hahlbohm2026faster}:
\begin{equation}
    S^{(k)} = R\left(\{I_i^{(3)}\}_{i=1}^M\right),
\end{equation}
where $S^{(k)}$ denotes the scene obtained from the $k$-th run. Novel views are rendered as
\begin{equation}
    V_j^{(k)} = \Pi_j\left(S^{(k)}\right), \quad j = 1, \ldots, T.
\end{equation}

In our final system, the scene is optimized for 30k iterations. Since repeated runs under the same setting may still produce slightly different local results, we average the rendered outputs from multiple independent runs,
\begin{equation}
    \bar{V}_j = \frac{1}{n} \sum_{k=1}^{n} V_j^{(k)}, \quad j = 1, \ldots, T,
\end{equation}
where $n=91$ in the final submission. This step is used to reduce unstable local artifacts.

\subsection{Implementation details}

The main implementation settings are summarized in Table~\ref{tab:implementation}. Additional scripts and implementation details are available in our public repository: \url{https://github.com/plbbl/GenSmoke-GS}.

\begin{table}[t]
    \centering
    \caption{Main implementation settings.}
    \label{tab:implementation}
    \begin{tabular}{lc}
        \toprule
        \textbf{Parameter} & \textbf{Value} \\
        \midrule
        MLLM model & GPT-Image-1.5 \\
        Optimization iterations & 30,000 \\
        Number of runs & 91 \\
        \bottomrule
    \end{tabular}
\end{table}

\begin{figure*}[t]
    \centering
    \includegraphics[width=0.98\textwidth]{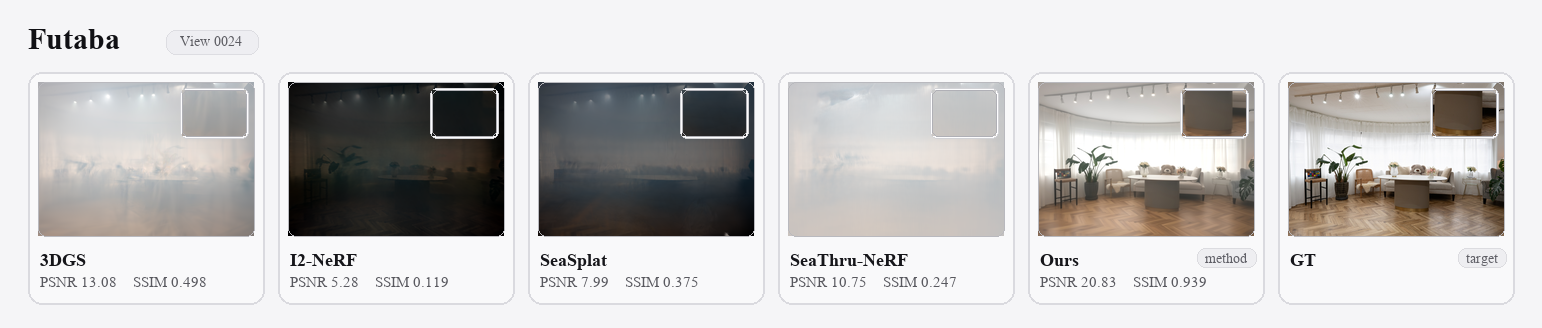}
    \caption{Representative qualitative comparison on Futaba, view 0024.}
    \label{fig:qualitative_main_1}
\end{figure*}

\begin{figure*}[t]
    \centering
    \includegraphics[width=0.98\textwidth]{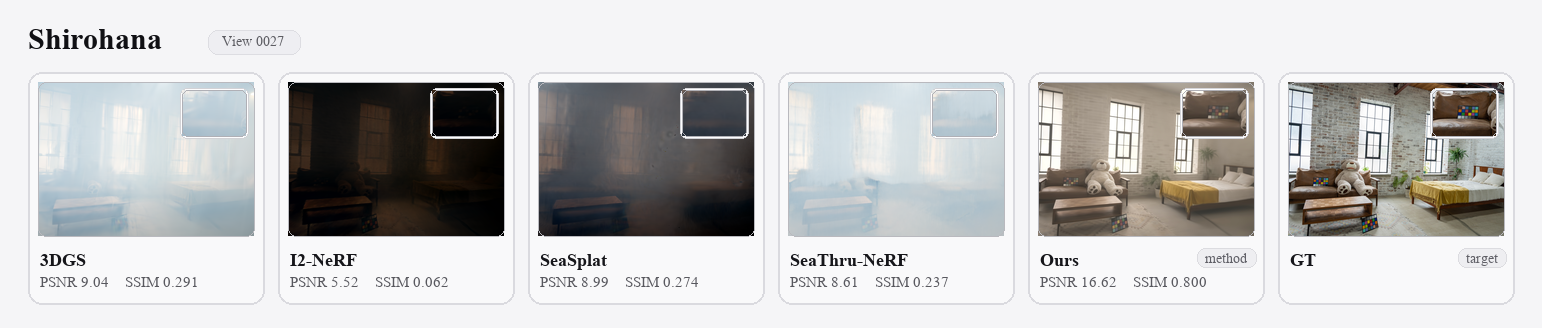}
    \caption{Representative qualitative comparison on Shirohana, view 0027.}
    \label{fig:qualitative_main_2}
\end{figure*}
\section{Experiment}
\label{subsec:experiment}

\subsection{Benchmark setting}

We evaluate the method on the RealX3D benchmark~\cite{liu2025realx3d} used in Track 2 of the NTIRE 2026 3D Restoration and Reconstruction (3DRR) Challenge. The benchmark contains smoke-degraded scenes and measures the quality of novel view synthesis. We report PSNR, SSIM, and LPIPS.

\subsection{Compared methods}

We compare the method with several available baseline outputs for degraded-scene rendering, including 3DGS~\cite{kerbl20233d}, I2-NeRF~\cite{liu2025i2nerf}, SeaSplat~\cite{yang2025seasplat}, and SeaThru-NeRF~\cite{levy2023seathru}. All quantitative results are computed from the provided rendered outputs and the corresponding test-set ground-truth views.

\subsection{Quantitative results}

Table~\ref{tab:main_results} reports the average results over all evaluated scenes. The method achieves the best performance on all three metrics. Compared with the strongest baseline 3DGS, it improves PSNR from 11.54 to 20.21 and SSIM from 0.597 to 0.729, while reducing LPIPS from 0.705 to 0.446.

\begin{table}[t]
    \centering
    \caption{Average quantitative results on the NTIRE 3DRR test set.}
    \label{tab:main_results}
    \setlength{\tabcolsep}{5pt}
    \begin{tabular}{lccc}
        \toprule
        \textbf{Method} & \textbf{PSNR$\uparrow$} & \textbf{SSIM$\uparrow$} & \textbf{LPIPS$\downarrow$} \\
        \midrule
        3DGS~\cite{kerbl20233d} & 11.54 & 0.597 & 0.705 \\
        I2-NeRF~\cite{liu2025i2nerf} & 7.13 & 0.257 & 0.852 \\
        SeaSplat~\cite{yang2025seasplat} & 9.00 & 0.440 & 0.827 \\
        SeaThru-NeRF~\cite{levy2023seathru} & 9.14 & 0.566 & 0.767 \\
        \midrule
        \textbf{GenSmoke-GS (Ours)} & \textbf{20.21} & \textbf{0.729} & \textbf{0.446} \\
        \bottomrule
    \end{tabular}
\end{table}

Table~\ref{tab:per_scene} further shows the per-scene performance. The improvements are consistent across the evaluated scenes, although some scenes remain difficult. For example, Shirohana still shows relatively lower results, indicating that severe degradation remains challenging.

\begin{table}[t]
    \centering
    \caption{Per-scene results of GenSmoke-GS.}
    \label{tab:per_scene}
    \setlength{\tabcolsep}{10pt}
    \begin{tabular}{lcc}
        \toprule
        \textbf{Scene} & \textbf{PSNR$\uparrow$} & \textbf{SSIM$\uparrow$} \\
        \midrule
        Futaba & 20.97 & 0.807 \\
        Hinoki & 19.12 & 0.607 \\
        Koharu & 20.87 & 0.783 \\
        Midori & 20.56 & 0.791 \\
        Natsume & 20.76 & 0.745 \\
        Shirohana & 17.50 & 0.569 \\
        Tsubaki & 21.66 & 0.799 \\
        \midrule
        \textbf{Average} & \textbf{20.21} & \textbf{0.729} \\
        \bottomrule
    \end{tabular}
\end{table}

\subsection{Qualitative results}

Figure~\ref{fig:qualitative_main_1} and Figure~\ref{fig:qualitative_main_2} show two representative qualitative comparisons from the test set. Each figure contains one scene and one target view. The compared methods are arranged from left to right as 3DGS, I2-NeRF, SeaSplat, SeaThru-NeRF, GenSmoke-GS, and ground truth.

\subsection{Discussion}

The results indicate that the method improves view synthesis quality under smoke degradation. The preprocessing and enhancement stages improve visibility before rendering, and repeated runs help reduce unstable local artifacts.

\section{Conclusion}
\label{sec:conclusion}

This paper presents GenSmoke-GS, a multi-stage method for novel view synthesis from smoke-degraded images in the NTIRE 2026 3DRR Challenge. The method combines image restoration, dehazing, MLLM-based enhancement, 3DGS-MCMC optimization, and averaging over repeated runs.

The experimental results show that this method is effective for the challenge setting. In particular, limiting structural changes during enhancement is helpful for stable optimization, and averaging repeated runs further improves the final rendered views.

\section*{Acknowledgments}
The authors thank the organizers of the NTIRE 2026 3D Restoration and Reconstruction Challenge for providing the benchmark and evaluation platform.

{
    \small
    \bibliographystyle{ieeenat_fullname}
    \bibliography{main}
}

\end{document}